\documentclass[10pt, a4paper, conference , twocolumn]{IEEEtran}
\usepackage{amsmath}
\usepackage{amssymb}
\usepackage{amsthm}
\usepackage{graphicx}
\usepackage{url}
\usepackage{array}
\usepackage[T1]{fontenc}
\usepackage[utf8]{inputenc}

\theoremstyle{definition}
\newtheorem{mydef}{Definition}

\usepackage{fancyhdr}
\usepackage[a4paper, hmargin=13mm, vmargin={27mm,45mm}, headsep=4mm]{geometry}

\begin{document}
\title{License Plate Recognition with Compressive Sensing Based Feature Extraction}

\author{
% First author - Prvi autor
\IEEEauthorblockN{Andrej Joki\'{c}}
\IEEEauthorblockA{
University of Montenegro\\
Podgorica, Montenegro \\
Email: aj.jokic@gmail.com}
% Second author - Drugi autor
\and
\IEEEauthorblockN{Nikola Vukovi\'{c}}
\IEEEauthorblockA{
University of Montenegro\\
Podgorica, Montenegro \\
Email: nikolav54@gmail.com}
% Other authors - Ostali autori
}
\maketitle \thispagestyle{fancy} % Do not change this - ne mijenjajte ovu liniju

\begin{abstract}
License plate recognition is the key component to many automatic traffic control systems. It enables the automatic identification of vehicles in many applications. Such systems must be able to identify vehicles from images taken in various conditions including low light, rain, snow, etc. In order to reduce the complexity and cost of the hardware required for such devices, the algorithm should be as efficient as possible. This paper proposes a license plate recognition system which uses a new approach based on compressive sensing techniques for dimensionality reduction and feature extraction. Dimensionality reduction will enable precise classification with less training data while demanding less computational power. Based on the extracted features, character recognition and classification is done by a Support Vector Machine classifier.
\end{abstract}

\begin{IEEEkeywords}
\textit{ compressive sensing; license plate recognition; support vector machine; TV minimization, optical character recognition }
\end{IEEEkeywords}

\section{Introduction}
In the last decade, the development and wide adoption of Intelligent Transportation Systems (ITS) has resulted in a growing demand for a reliable license plate recognition system. Examples of applications of such systems include speeding radars, surveillance systems, automatic parking access systems etc. \cite{lpr,feat}

At the base of any license plate recognition system is \emph{Optical Character Recognition} (OCR). Most OCR systems consist of three steps. First, the image must be analyzed by splitting it into small segments that contain one character each. This step is called \emph{segmentation} \cite{lpr}.

In order to recognize characters from images, a \emph{support vector machine} (SVM) classifier will be used. SVM is a statistical classifier based on supervised learning. It is able to recognize certain patterns by analyzing labelled training data and to classify the data based on these patterns. However, raw images contain a large amount of data that is useless to the classifier, thereby reducing the efficiency and requiring more training data. Because of this, instead of analyzing raw data, classifiers are supplied with a small number of features that describe the data. The process of selecting and calculating these features is called \emph{feature extraction} \cite{feat}.

In this paper, a method of feature extraction based on the \emph{compressive sensing} theory will be presented. Compressive sensing is a fairly new field that has been the subject of intensive research in the last decade. This new method of sampling enables the acquisition of data with far fewer samples than required by the \emph{Shannon-Nyquist sampling theorem}. If the original data and the sampling method meet certain conditions, the data can be fully reconstructed from this small number of samples by solving certain mathematical optimization problems. Based on this, we can deduce that these samples must contain enough information about these images to recognize them. The fact that they contain a lot of information in a small amount of data makes these samples a perfect candidate for classification features \cite{feat_cs,feat_cs2}.

\section{Preprocessing}
The first step in the recognition of the characters on a license plate is \emph{segmentation}. This process involves dividing the image into smaller segments. Each segment should contain one character of the license plate text \cite{lpr}. 

This task is usually performed by a simple algorithm. First, an adaptive threshold is applied in order to obtain a binary image. This simplifies the next steps and eliminates the variation of features due to different brightness. The image is then analyzed and all the connected objects are extracted into separate images. Naturally, this simple algorithm cannot distinguish between characters and other objects, but this is the task of the classifier. In order to reduce the number of segments being analyzed, segments containing less than a certain number of pixels are discarded immediately \cite{lpr}.

\begin{figure}[h]
\centering
\includegraphics[width=0.3\textwidth]{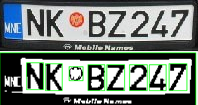}
\caption{Original and segmented license plate image}
\label{plt_seg}
\end{figure}

As figure \ref{plt_seg} shows, besides the license plate number and the text above it, the algorithm also detects other objects such as the frame of the license plate. Smaller objects are successfully ignored by the algorithm.

\section{Feature extraction}
\label{sec:extr}

Feature extraction is the single most important step in character recognition. Features enable the classifier to recognize a character and distinguish it from other characters. Because of this, choosing the appropriate features directly affects the performance and precision of the classifier. 

In order to ensure precise distinction between different characters, the extracted features should satisfy a few conditions. Firstly, the features should be invariant to the expected distorsions and variations that a character may have in a specific image. For example, the size of the character should not affect any of the features. Because of this, the segments from the previous step should be scaled to a fixed size. The binarization which is also done in the segmentation step ensures that the brightness and color of the characters does not affect the features. In other words, the features for different images of the same character should be as similar as possible \cite{feat}. 

Another condition that must be satisfied is the size of the feature vectors. Namely, larger numbers of features can negatively affect classification performance and require more training data. Because of this, features should represent only the information that will be useful to make a distinction between various characters \cite{feat}. 

\subsection{Compressive sensing}

In order to reduce the number of features, this paper proposes a \emph{dimensionality reduction} method based on compressive sensing. Compressive sensing is a concept proposed as an improvement to the \emph{Shannon-Nyquist sampling theorem} which is the fundamental theorem of signal processing. The Shannon-Nyquist sampling theorem states that a continuous signal can be reconstructed from a digital signal whose sampling frequency is at least twice as high as the maximal frequency of the signal. This results in a large amount of data, most of which will be discarded when the signal is compressed. Compressive sensing overcomes this problem by directly taking a small number of measurements from the signal \cite{mms}.

In mathematical terms, the measured data $y \in \mathbb{C}^{m}$ is related to the original signal $x \in \mathbb{C}^{N}$ via \cite{mms,cs_knjiga}

\begin{equation}
\label{eq:meas}
\mathbf{A}x = y.
\end{equation}

The matrix $\mathbf{A} \in \mathbb{C}^{m\times N}$ models the linear measurement process. In order to reconstruct the original signal, the above linear system has to be solved. According to the Shannon-Nyquist sampling theorem, $m$ must be at least as large as $N$, and (\ref{eq:meas}) yields an unique solution. In the case that $m<N$, classical linear algebra indicates that the linear system (\ref{eq:meas}) is undetermined and has an infinite number of solutions, making recovery of the original signal $x$ impossible. However, according to compressive sensing, it is actually possible to reconstruct $x$ even from $m\ll N$ measurements if certain conditions are met \cite{mms,dsp}. 

One of the conditions that must be met is sparsity, i.e. the signal must be sparse in a certain transformation domain in order to ensure a successful reconstruction. The signal is assumed to be $s$-sparse in one of the common transformation domains. 

An important factor that affects the signal reconstruction is the quality of the measurement matrix. One good metric for the quality of a measurement matrix is the \emph{restricted isometry property} (RIP). This concept was introduced in \cite{rip}. 

\begin{mydef}
\cite{cs_knjiga} The $s$-th \emph{restricted isometry constant} $\delta_{s}=\delta_{s}(\mathbf{A})$ (depending on $s$ and $\mathbf{A}$) of a matrix $\mathbf{A} \in \mathbb{C}^{m\times N}$ is the smallest $\delta \ge 0$ such that
\begin{equation}
(1-\delta)\left\|x\right\|^{2}_{2} \le \left\|\mathbf{A}x\right\|^{2}_{2}\le(1+\delta)\left\|x\right\|^{2}_{2}
\end{equation}
for all $s$-sparse vectors $x\in\mathbb{C}^{N}$.
\end{mydef}

We say that $\mathbf{A}$ satisfies the retricted isometry property if $\delta_{s}$ is small for a reasonably large $s$. 

% Another very important notion concerning the measurement matrix is \emph{coherence}. Generally, a smaller coherence (i.e. a \emph{incoherence} condition) yields better reconstruction results. This notion is defined in the following definition.

% \begin{mydef}
% \cite{cs_knjiga} Let $\mathbf{A}\in\mathbb{C}^{m\times{}N}$ be a matrix with $l_{2}$-normalized columns $a_{1},...,a_{N}$, i.e. $\left\|a_{i}\right\|_{2}=1$ for all $i\in [N]$. The \emph{coherence} $\mu=\mu(\mathbf{A})$ of the matrix $\mathbf{A}$ is defined as

% \begin{equation}
% \label{eq:coherence}
% \mu := \max_{1\le i \ne j \le N} |\langle a_{i},a_{j}\rangle|.
% \end{equation}
% \end{mydef}

Some of the commonly used measurement matrices that satisfy this condition are random Gaussian, Bernoulli and partial random Fourier matrices. In this paper, a random Bernoulli matrix will be used. The lower bound for the number of measurements with this matrix is \cite{mms}:

\begin{equation}
m\ge C\cdot s\cdot \log\frac{N}{s}
\end{equation}

The development of a resonably fast algorithm for signal reconstruction is very important. The first algorithmic approach that comes to mind is $l_{0}$ minimization, i.e. the search for the sparsest vector consistent with the measured data (the $l_0$-norm of a vector $z\in\mathbb{C}^N$ is defined as a number of its non-zero components). However, this is generally a very NP-hard problem and it is therefore not viable for use in practice. A very popular alternative method is $l_{1}$ minimization, also known as \emph{basis pursuit}, which consists in finding the minimizer of the following problem \cite{mms,dsp,cs_knjiga}:

\begin{equation}
\text{minimize} \left\|z\right\|_{1} \quad\text{subject~to}\quad \mathbf{A}z=y,
\end{equation}

This optimization problem can be solved with efficient methods from convex optimization. Basis pursuit can be interpreted as the convex relaxation of $l_{0}$ minimization. Alternative methods include greedy-type methods such as orthogonal matching pursuit, as well as thresholding-based methods including iterative hard thresholding \cite{cs_knjiga}. 

In this paper, the \emph{total variation} (TV) method will be used in order to test the reconstructibility of the character images from the measured samples. This method provides great results in image processing applications. It is based on solving the following optimization problem \cite{mms,a18}:

\begin{equation}
\min_{x} TV(x) \quad\text{subject to}\quad \left\|\mathbf{A}x-y\right\|_{2}^{2}<\varepsilon^{2}
\end{equation}

The total variation of $x$ represents the sum of the gradient magnitudes at each point. %and can be approximated as:
% \begin{equation}
% TV(x)=\sum_{i,j}{\left\|D_{i,j}x\right\|_{2}},\; D_{i,j}x=\left[
% \begin{aligned}
% x(i+1,j)-x(i,j)\\
% x(i,j+1)-x(i,j)
% \end{aligned}
% \right].
% \end{equation}

Since a perfect reconstruction of the original signal from the compressive measurements is possible, we can assume that these measurements contain descriptive information about the signal. Therefore, these samples can be used as features in a classification algorithm \cite{feat_cs,feat_cs2}. In the proposed approach, we will take compressed measurements from the character images using binary measurement matrices. The measurement vectors will then be used as feature vectors for training and testing the classifier.

\section{The proposed approach}

The previous chapters have given a brief summary of the theory on which this paper is based. We will now summarize an algorithm that implements the proposed approach. %A flowchart of the algorithm is given in Figure \ref{alg} (also see \cite{lpr2}). 

First, the image of the license plate is converted to a black and white image. All groups of connected pixels are then extracted into separate images. In order to make the algorithm insensitive to image size, all the images are scaled to the same size. This completes the \emph{segmentation} step \cite{lpr}. 

Each character image (i.e. segment) is then multiplied by the measurement matrix. This way, a small number of random samples will be collected from the signal. According to the CS theory from the previous chapter, these samples contain enough information about the signal to fully reconstruct it, and can therefore be used as features \cite{feat_cs,feat_cs2}. %The samples are placed into a vector and used as feature vectors.

In order to determine what character does an image represent, a support vector machine (SVM) classifier is used. Since we have multiple classes, we have to use a multiclass SVM which is a modified version of the standard classifier \cite{lpr}. The training data for the classifier is a large set of different images and their corresponding labels. When the classifier is trained it creates a model that will be used to determine which class does an image belong to. In order to test the precision of the classifier, the available data is randomly split into a training set and a testing set. After training, the model's accuracy is measured by applying it to the testing set and comparing the predicted labels with the original labels. 

% \begin{figure}[h]
% \centering
% \includegraphics[width=0.3\textwidth]{alg.pdf}
% \caption{Flowchart of the proposed approach}
% \label{alg}
% \end{figure}

\section{Results}

In order to verify the performance of the proposed system, we have done a simulation on a sample character image set. The simulation has been done in \emph{Matlab 9.3}.

The test was done using a character image set available in the Computer Vision System toolbox for Matlab. To reduce the complexity of the simulation, the testing was done using only digits (0-9). The aforementioned image set contains 1010 images of digits, i.e. 101 image for each digit, with dimensions of $16 \times 16$ pixels. A commonly used split ratio in machine learning 4 by 1, i.e. 80\% is used for training and 20\% for testing. To get more consistent results, the test was run multiple times (20), and the obtained results were averaged.

\begin{figure}[h]
\centering
\includegraphics[width=0.35\textwidth]{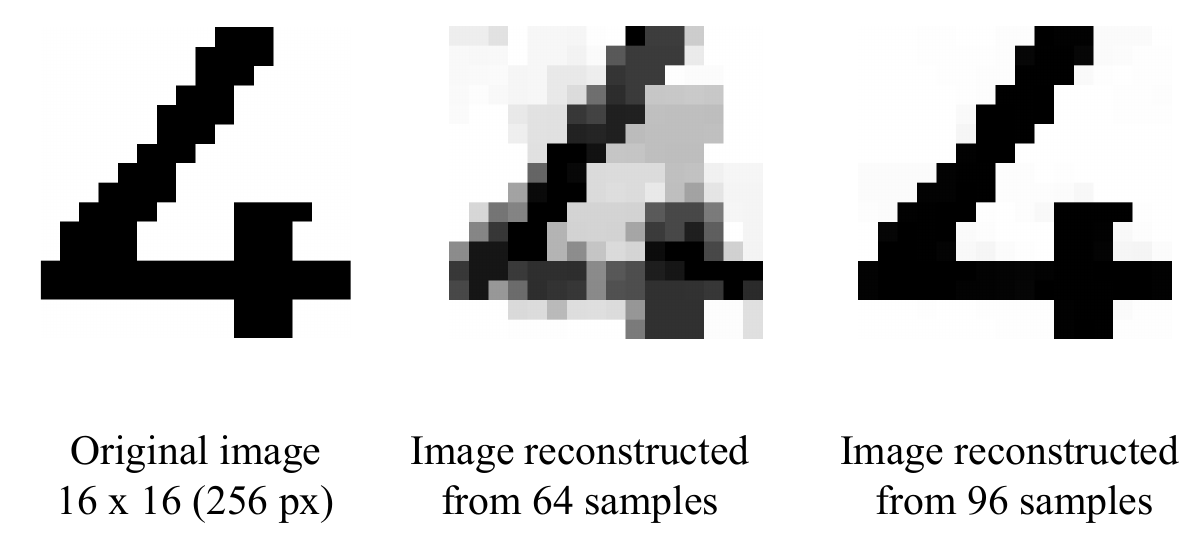}
\caption{Character image reconstructed from different numbers of compressed samples}
\label{reconstr}
\end{figure}

The images were sampled using a random Bernoulli matrix in which each element takes values $\pm1$ with $1/2$ probability. Figure \ref{reconstr} shows a sample character image and its reconstructions from different numbers of samples. The reconstruction has been done using the $TV$ minimization algorithm from the $l_1$-\emph{magic} toolbox \cite{magic}

The samples extracted from the images have been used as features for the classification step. The classifier used in this test was a multiclass SVM with error correcting output codes, using a 'one-vs-one' coding scheme. Firstly, the training set of features and labels was used to train a model. Then, the testing set of features was applied to the model in order to test the classification accuracy. 

In the first test, 64 features were extracted from the character. By running the test 20 times (with different random measurement matrices and data splits), an average total accuracy of $98.81\%$ has been achieved. The minimum total accuracy that was observed was $97.52\%$, while the maximum was $100\%$. In each run, a confusion matrix was generated, and the average scores for each character are shown in Table \ref{tab:t1}.

\begin{table}[tbh]
	\caption{Average classification scores for each character from 64 samples}
	\label{tab:t1}
	\centering
	\begin{tabular}{>{\bfseries}c | c c c}
	\small
	Chr & \textbf{Precision} & \textbf{Recall} & \textbf{F1 Score} \\
	\hline
	0 &   98.233   &    99.424 &   98.778  \\
    1 &   97.993   &    99.808 &   98.855  \\
    2 &   98.662   &    98.028 &   98.257  \\
    3 &   98.811   &    97.982 &   98.363  \\
    4 &      100   &    99.273 &   99.621  \\
    5 &   97.924   &    98.546 &   98.156  \\
    6 &   98.318   &    98.491 &   98.332  \\
    7 &     99.5   &    98.856 &    99.13  \\
    8 &   99.655   &    98.415 &   99.002  \\
    9 &    99.47   &    98.854 &   99.128  \\
	\end{tabular}
\end{table}

Another test was done with 96 features. This time, the results were slightly better, achieving an average of $99.08\%$ and a minimum of $98.02\%$. The scores by character are given in Table \ref{tab:t2}.

\begin{table}[tbh]
	\caption{Average classification scores for each character from 96 samples}
	\label{tab:t2}
	\centering
	\begin{tabular}{>{\bfseries}c | c c c}
	\small
	Chr & \textbf{Precision} & \textbf{Recall} & \textbf{F1 Score} \\
	\hline
	0 &      100  &     99.412 &   99.688 \\  
    1 &   97.582  &       99.5 &   98.449 \\
    2 &   98.689  &     98.699 &   98.642 \\
    3 &   99.231  &      98.74 &   98.922 \\
    4 &      100  &        100 &      100 \\
    5 &    98.57  &      98.19 &   98.308 \\
    6 &    98.53  &     99.093 &   98.764 \\
    7 &      100  &        100 &      100 \\
    8 &      100  &     97.814 &   98.853 \\
    9 &   99.129  &     99.583 &   99.327 \\
	\end{tabular}
\end{table}

As Figure \ref{reconstr} shows, 64 samples are not enough for a decent image reconstruction. However, the classification results show that this amount of data still provides a fairly accurate recognition. A classification test has been done with as low as 32 samples, still achieving an average of $98.2\%$ accuracy, while a TV reconstruction from these samples was practically unrecognizable. This shows that the proposed method of classification is very robust and can provide effective recognition even from a very small number of measurements.

\section{Conclusion}

The intensive research in comressive sensing reveals many new possible applications of the theory. In this paper, yet another successful application of these techniques has been demonstrated. The results show that compressive sensing based feature extraction has great performance in the classification of character images. The simple measurement matrix gives this method an advantage over other popular methods in terms of computational complexity.

\end{document}